\documentclass{article}
\usepackage{spconf,amsmath,graphicx}
\usepackage{subfig}
\usepackage[justification=centering]{caption}
\usepackage{bm}
\usepackage{amsfonts}

\title{EVERY FILTER EXTRACTS A SPECIFIC TEXTURE IN CONVOLUTIONAL NEURAL NETWORKS}
\name{Zhiqiang Xia \quad Ce Zhu \quad Zhengtao Wang \quad  Qi Guo \quad Yipeng Liu}
\address{School of Electronic Engineering / Center for Robotics\\
		University of Electronic Science and Technology of China (UESTC), Chengdu, China\\
		Email: \{eczhu, yipengliu\}@uestc.edu.cn, \{zhiqiangxia\}@std.uestc.edu.cn}
\begin{document}
\ninept
\maketitle
\begin{abstract}
Many works have concentrated on visualizing and understanding the inner mechanism of convolutional neural networks (CNNs) by generating images that activate some specific neurons, which is called deep visualization. However, it is still unclear what the filters extract from images intuitively. In this paper, we propose a modified code inversion algorithm, called feature map inversion, to understand the function of filter of interest in CNNs. We reveal that every filter extracts a specific texture. The texture from higher layer contains more colours and more intricate structures. We also demonstrate that style of images could be a combination of these texture primitives. Two methods are proposed to reallocate energy distribution of feature maps randomly and purposefully. Then, we inverse the modified code and generate images of diverse styles. With these results, we provide an explanation about why Gram matrix of feature maps \cite{Gatys_2016_CVPR} could represent image style.\\
\end{abstract}
\section{INTRODUCTION}
\label{sec:intro}
Convolutional neural networks (CNNs) have achieved tremendous progress on many pattern recognition tasks, especially large-scale images recognition problems \cite{szegedy2015going,krizhevsky2012imagenet,Simonyan14c,He_2016_CVPR}. However, on one hand, CNNs still make mistakes easily. \cite{goodfellow2014explaining} reveals that adding adversary noise to an image in a way imperceptible to humans can cause CNNs mislabel the image. \cite{nguyen2015deep} shows some related results: it is easy to use evolutionary algorithms to generate images that are completely unrecognizable to humans, but the state-of-the-art CNNs can classify the image to some categories with 99.99\% confidence. On the other hand, it is still unclear how CNNs learn suitable features from the training data and what a feature map represents \cite{yosinski2014transferable,li2015convergent}. This dimness of CNNs promote recent development of visualization of CNNs, also known as deep visualization \cite{erhan2009visualizing,szegedy2013intriguing,springenberg2014striving,wei2015understanding,lenc2015understanding,karpathy2015visualizing,liu2016towards}. Deep visualization aims to reveal the internal mechanism of CNNs by generating an image that activates some specific neurons, which could provide meaningful and helpful insights for designing more effective architectures \cite{shang2016understanding,zeiler2014visualizing}.

There are many deep visualization techniques available for understanding CNNs. Perhaps the simplest way is displaying the response of a specific layer, or several special feature maps \footnote{To avoid the conception confusion, in this paper, ``code'' is adopted to represent the activations of a whole layer, and ``feature map'' is utilized to represent activation of a single filter at a layer.}. However, these feature maps only provide limited and unintuitive information about filters and images. For instance, although it is possible to find some filters that respond to a specific object such as “face” in \cite{sun2015deeply}, this method is heuristic and not universal.

A major deep visualization technique is activation maximization \cite{erhan2009visualizing}, which finds an image that activates some specific neurons most intensively to reveal what feature these neurons response to. \cite{simonyan2013deep} shows the object conception learned by AlexNet by maximizing the neuron activation at the last layer. \cite{yosinski2015DL} generates similar results by applying the activation maximization to a single feature map. \cite{mordvintsev2015inceptionism} generates many fantastic images by intensifying the activated neurons of input images at high layers as well as low layers, which is called ``deep dream''. However, the generated images are rough. So a series of subsequent works concentrated on improving generated images quality by adding natural priors, such as $\ell_2$ norm \cite{simonyan2013deep}, total variation \cite{Mahendran2016}, jitter \cite{mordvintsev2015inceptionism}, Gaussian blur \cite{yosinski2015DL} and data-driven patch priors \cite{wei2015understanding}. Besides, \cite{nguyen2016multifaceted} also uncovers the different types of features learned by each neuron with a priori input image.

Another major deep visualization technique calls code inversion \cite{Mahendran2016}, which generates an image whose activation code is similar to the target activation code at a particular layer produced by a specific image. It reveals which features are extracted by filters from the input image. Code inversion could be realized by training another neural network and directly predicting the reconstructed image \cite{dosovitskiy2015inverting}, or by iteratively optimizing an initial noisy image \cite{7299155}, or transposing CNNs to project the feature activations back to the input pixel space with deconvnet \cite{zeiler2014visualizing}. These {}inversion methods could also be extended to statistical property of code. \cite{NIPS2015_5633} visualizes Gram matrix of feature maps and finds that it represents image texture. \cite{Gatys_2016_CVPR,ulyanov2016texture,johnson2016perceptual} utilize Gram matrix to do image style transfer. Compared with activation maximization, code inversion could intuitively reveal the specific feature extracted by filters from given images.

Many previous works in deep visualization have revealed some valuable explanations about a single neuron \cite{simonyan2013deep,nguyen2015deep}, a feature map \cite{yosinski2015DL} or the code \cite{mordvintsev2015inceptionism} at different layers. The CNNs are not totally black boxes anymore. However, to our best knowledge, there is still no work about visualizing what exactly every filter tries to capture from an image. Understanding of filters could help improve architecture of CNNs.

In this paper, we introduce \emph{Feature Map Inversion} (FMI) to deal with the aforementioned problem. For a filter of interest, FMI enhances the corresponding feature map and weakens the rest feature maps at the same time. Then classical code inversion algorithm is applied to the modified code and generate inversion images. Our experimental results show that every filter in CNNs extracts a specific texture. The texture at higher layers contains more colours and more intricate structures (Fig. \ref{fig:feature_map}). In addition, we find that style of an image could be a combination of hierarchical texture primitives. Two methods are proposed to generate images of diverse styles by inversing modified code. In particular, we change code by reallocating the sum of each feature map randomly, and according to target code purposefully. With these results, we provide an explanation about why Gram matrix of feature maps \cite{Gatys_2016_CVPR} could represent image style. Since every filter extracts a specific texture, the combination weights of feature maps decides image style. Like the sum of feature maps along channel axis, Gram matrix also guides the energy of every feature map of generated image.

Our experiments were conducted based on the open-source deep learning framework Mxnet \cite{chen2015mxnet} and is available at https://github.co-\\m/xzqjack/FeatureMapInversion.

\section{METHOD}
\label{sec:method}
\subsection{Feature Map Inversion}
\label{sec:method_FMI}
\begin{figure}[tbp]
	\centering
	\includegraphics[width=240pt, keepaspectratio]{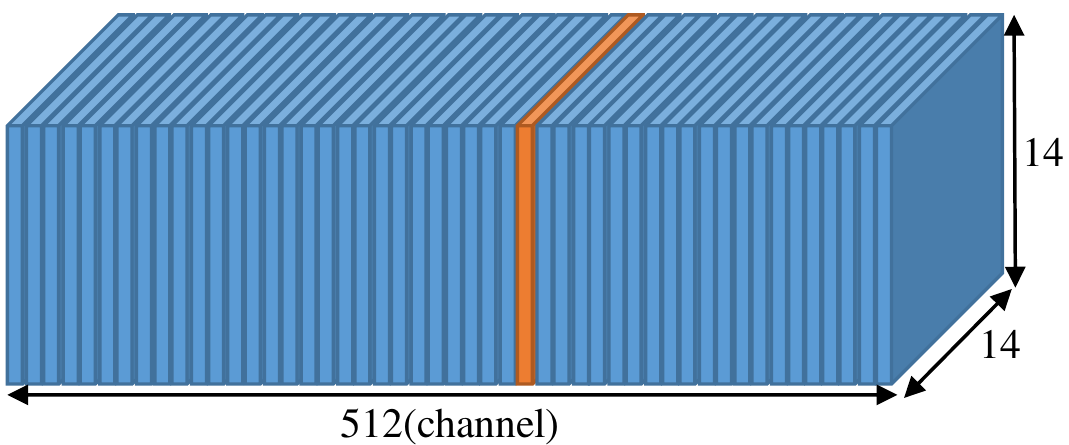}
	\caption{Modification of feature maps at layer relu5\_1 in VGG-19.}
	\label{fig:featue_maps-relu5_1}
\end{figure}

In this section, we use FMI to answer the question: ``What does a filter in CNNs try to capture from an input image''. Given an input image $\bm{X}\in \mathbb{R}^{3 \times H \times W}$, a trained CNN $\Phi$ encodes the image as code $\Phi(\bm{X})\in \mathbb{R}^{C \times M \times N}$. Code inversion method aims to find a new image $\bm{X}^*$ whose code $\Phi(\bm{X}^*)$ is similar to $\Phi(\bm{X})$. As shown in Fig. \ref{fig:featue_maps-relu5_1}, for a chosen layer such as relu5\_1 in VGG-19 \cite{Simonyan14c}, the code is a 3D tensor. There are totally 512 feature maps and the shape of each feature map is $14\times14$. In order to visualize what feature the $\emph{l}^{th}$ filter of interest extract, we should intensify the $\emph{l}^{th}$ feature map to a certain degree and weaken the others. In this paper, we set the $\emph{l}^{th}$ feature map to the sum of feature maps along channel axis and the others to 0. Finally, we apply classical code inversion \cite{Mahendran2016} to the modified code $\Psi(\Phi(\bm{X}),\emph{l})$.

\begin{equation}
\min_{\bm{X}^* \in \mathbb{R}^{3\times H \times W}} \lVert\Phi(\bm{X}^*)-\Psi(\Phi(\bm{X},\emph{l}))\rVert_F^2
\end{equation}

\begin{equation}
	\Psi(\Phi(\bm{X}),\emph{l})_{\emph{c},\emph{m},\emph{n}}=\begin{cases}
		\sum\limits_{k=1}^C{\Phi(\bm{X})}_{k,m,n}, &\mbox{\emph{c}=\emph{l}} \\		
		    \quad\quad\quad0\qquad\:\:,&\mbox{\textmd{otherwise}}
	\end{cases}
\end{equation}
where ${\Phi(\bm{X})}_{k,m,n}$ denotes the $k^{th}$ feature map activation at location $(m,n)$. $\Psi(\Phi(\bm{X}),\emph{l})$ enhances the $\emph{l}^{th}$ feature map of $\Phi(\bm{X})$ and weaken the others.

\subsection{Modified Code Inversion}
\label{sec:method_MCI}

Knowing that every filter extracts a specific texture, we assume that style of an image could be considered as the combination of hierarchical texture primitives. If so, we can combine the texture primitives by modifying the energy distribution of feature maps randomly and purposefully. Then, if we apply code inversion to the modified code, we will generate diverse style images.

The random modified method keeps the activation states (activated or not) of neurons unchanged and reallocate the sum of every feature map. We firstly generate a random vector $\bm{v}\in \mathbb{R}^C$, where $\sum_{c=1}^Cv_c=1$ and $v_c\geq0$. Then we reallocate the energy of every feature map with vector $\bm{v}$. The modified code is

\begin{equation}
\Psi(\Phi(\bm{X}))_{c,i,j}=\frac{\bm{v}_c\Phi(\bm{X})_{c,i,j}} {\sum\limits_{m=1}^M\sum\limits_{n=1}^N\Phi(\bm{X})_{c,m,n}}\sum\limits_{k=1}^C \sum\limits_{m=1}^M \sum\limits_{n=1}^N \Phi(\bm{X})_{k,m,n}
\end{equation}

We take the modified code as target and generate an image by

\begin{equation}
\min_{\bm{X}^* \in \mathbb{R}^{3\times H \times W}} \lVert\Phi(\bm{X}^*)-\Psi(\Phi(\bm{X}))\rVert_F^2
\end{equation}
The content of generated image keeps unchanged coarsely but the styles are variegated as shown in Fig. \ref{fig:random}.

One more step, we modify the energy of each feature map by making the sum of every feature map similar to the target code purposefully. In particular, suppose that we have two input images: content image $\bm{X}^\textmd{c} \in \mathbb{R}^{3 \times H^\textmd{c} \times W^\textmd{c}}$ and style image $\bm{X}^\textmd{s} \in \mathbb{R}^{3 \times H^\textmd{s} \times W^\textmd{s}}$. The feature maps of $\bm{X}^\textmd{c}$ and $\bm{X}^\textmd{s}$ at a layer are reshaped as $\Phi(\bm{X}^\textmd{c})\in \mathbb{R}^{C \times M^\textmd{c}N^\textmd{c}}$ and $\Phi(\bm{X}^\textmd{s})\in \mathbb{R}^{C \times M^\textmd{s}N^\textmd{s}}$. We use content code $\Phi(\bm{X}^\textmd{c})$ as content constraint \cite{Gatys_2016_CVPR} and sum of feature maps $\Phi(\bm{X}^\textmd{s})$ along channel axis as style constraint. Finally, we generate a styled image by
\begin{equation}
\begin{split}
\min_{\bm{X}^* \in \mathbb{R}^{3 \times H^\textmd{c} \times W^\textmd{c}}} {} &\alpha \lVert\Phi(\bm{X}^*)-\Phi(\bm{X}^\textmd{c})\rVert_F^2+{} \\						  &\beta\lVert\Phi(\bm{X}^*)\bm{I}^\textmd{c}- \Phi(\bm{X}^\textmd{s})\bm{I}^s\rVert_F^2
\end{split}
\end{equation}
where $\alpha$, $\beta$ are different weights of content term and style term. $\bm{I}^\textmd{c}\in \mathbb{R}^{M^\textmd{c}N^\textmd{c} \times 1}$, $\bm{I}^\textmd{s}\in \mathbb{R}^{M^\textmd{s}N^\textmd{s} \times 1}$ and all the values of $\bm{I}^\textmd{c}$ and $\bm{I}^\textmd{s}$ are $\bm{1}$.

\section{Experiment}
\label{sec:experiment}

\subsection{Experiment Setting}
\label{sec:setting}
\begin{figure}[tbp]
\centering
	\includegraphics[width=120pt, keepaspectratio]{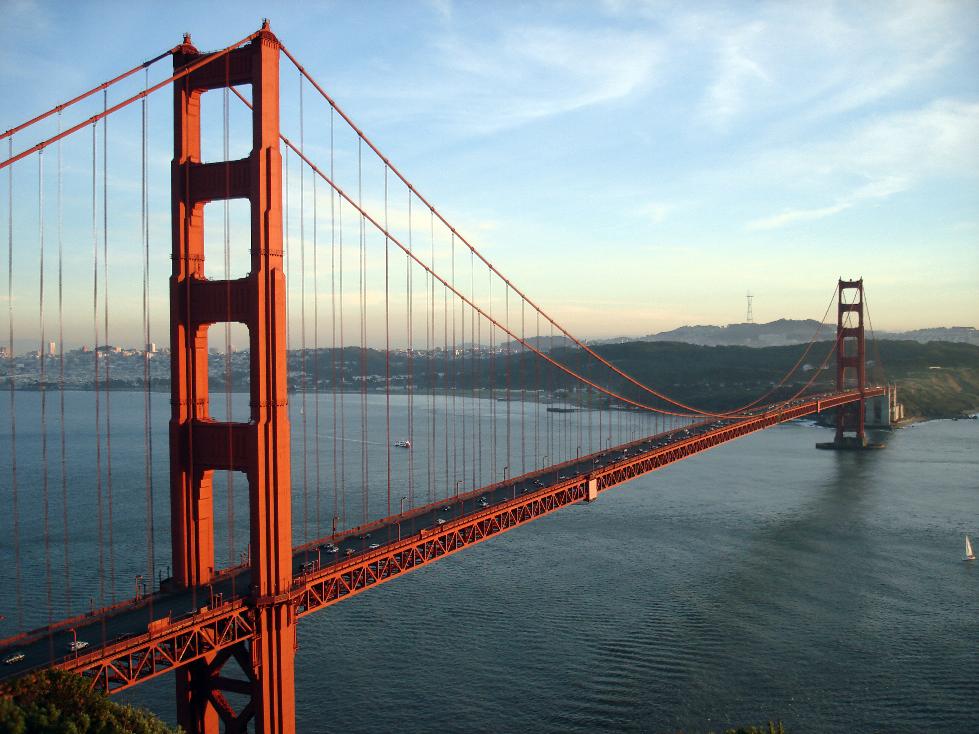}
	\includegraphics[width=120pt, keepaspectratio]{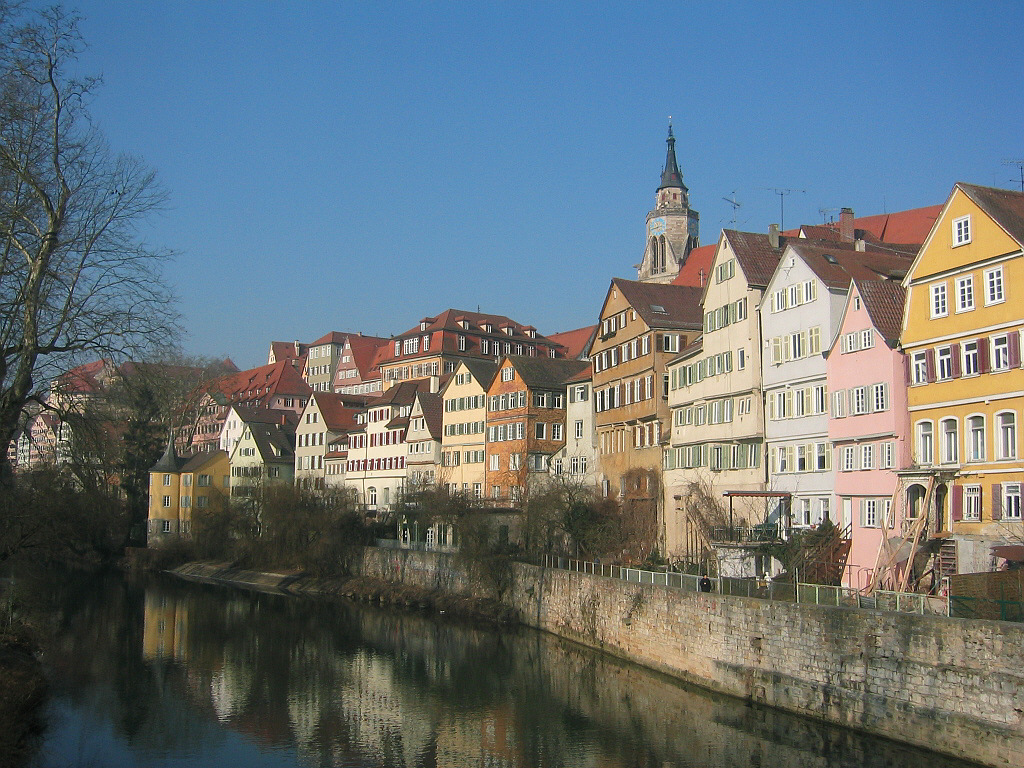}
	\caption{Content images. The left is \emph{The Golden Gate Bridge} taken by Rich Niewiroski Jr in 2007. The right is \emph{Neckarfront with Hölderlinturm and collegiate} taken by Andreas Praefcke in 2003.}
	\label{fig:content_image}
\end{figure}

We conduct our experiments on a well-known deep convolutional neuron networks, which is called VGG-19 \cite{Simonyan14c}. It is trained to recognize 1000 different objects on 1.2-million-image ILSVRC 2014 ImageNet dataset \cite{ILSVRC15}. It contains 16 convolutional layers, 16 relu layers, 5 pooling layers and totally 5504 filters. The size of all filters is $3 \times 3$. We do not use any fully connected layers. In the experiments, we set the $\alpha=10$, $\beta=1$ and use \emph{The Golden Gate Bridge} and \emph{Neckarfront with Hölderlinturm and collegiate} as content images in all cases as shown in Fig. \ref{fig:content_image}.

\subsection{Feature Map Inversion} 
\label{sec:result_FMI}
\begin{figure}[tbp]
	\centering
	\subfloat[\emph{The Golden Gate Bridge}]{
		\centering
		\label{fig:feature_map_Golden}
		\includegraphics[width=240pt, keepaspectratio]{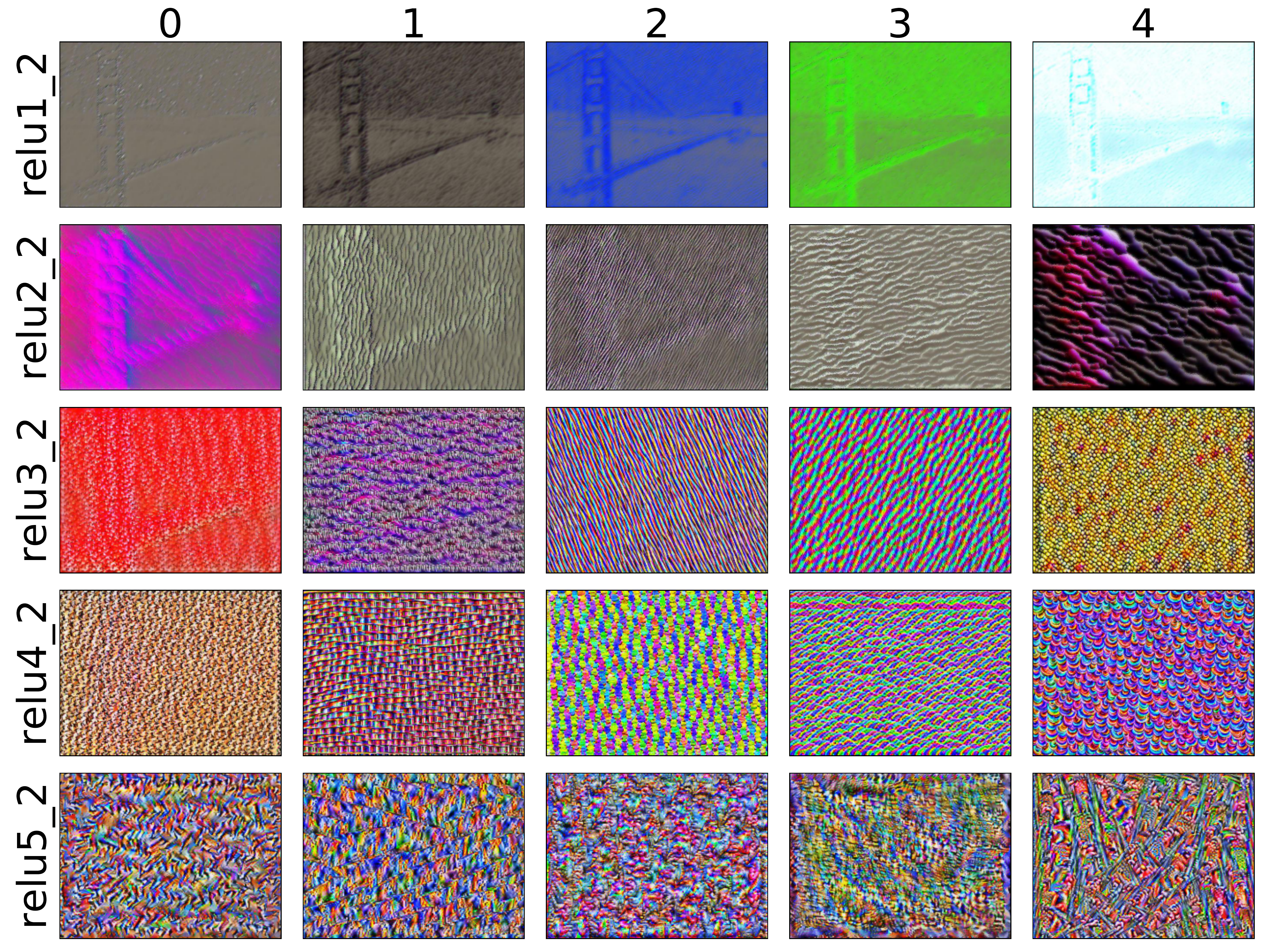}
	}
	\centering

	\subfloat[Neckarfront with Hölderlinturm and Collegiate]{
		\centering
		\label{fig:feature_map_tubingen}
		\includegraphics[width=240pt, keepaspectratio]{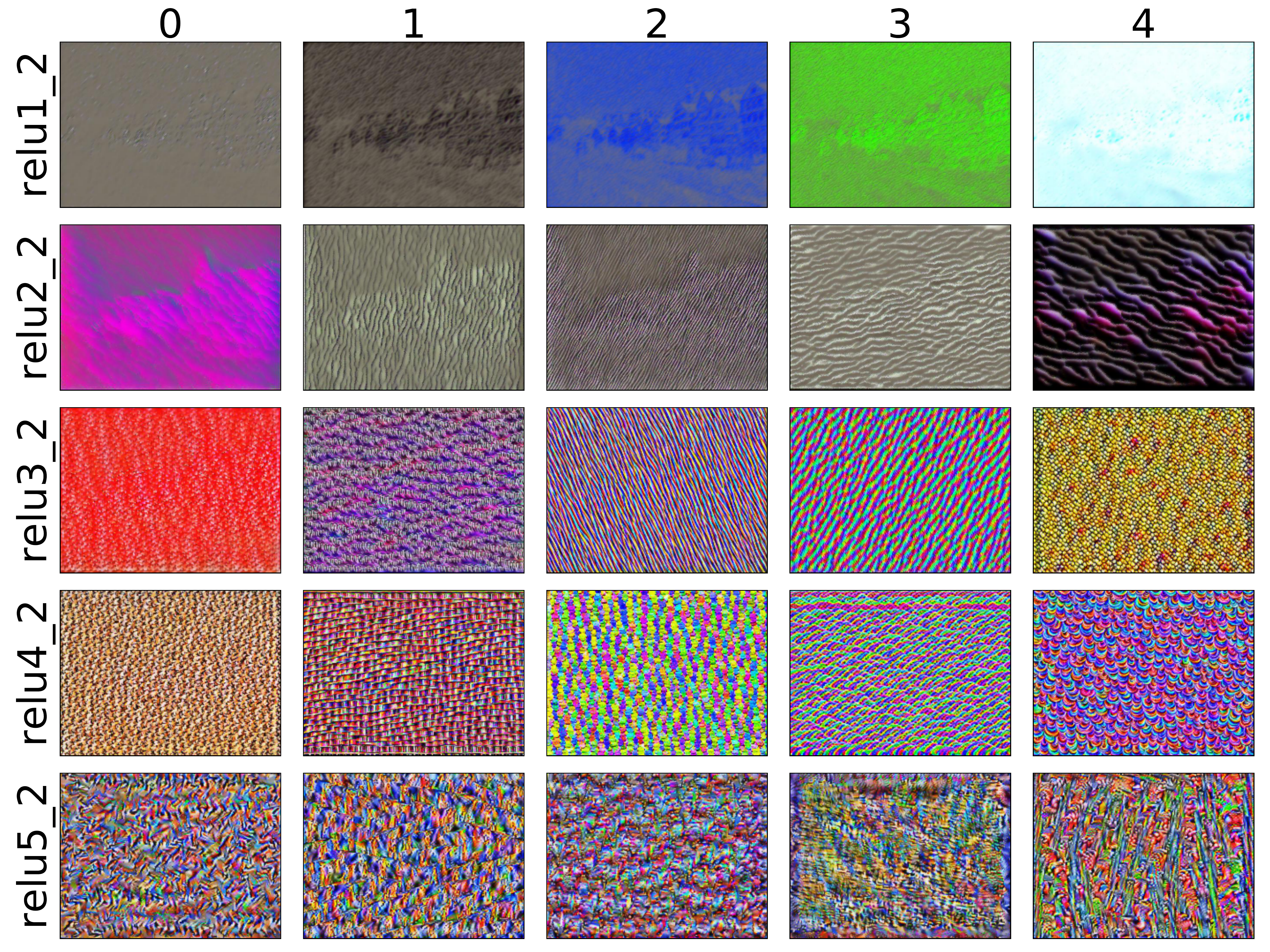}
	}
	\caption{Feature Map Inversion. The top inversion results is with input image \emph{The Golden Gate Bridge} and the bottom is with \emph{Neckarfront with Hölderlinturm and Collegiate}. The columns show the $1^{th}$, $2^{th}$, $3^{th}$, $4^{th}$ and $5^{th}$ MFI results respectively. The rows show FMI results at layers, relu1\_2, relu2\_2,relu3\_2, relu4\_2 and relu5\_2 respectively. This figure is best viewed electronically in color with zoom.}
	\label{fig:feature_map}
\end{figure}

We show qualitative FMI results on Fig. \ref{fig:feature_map}. The top inversion results are from input image \emph{The Golden Gate Bridge} and the bottom are from \emph{Neckarfront with Hölderlinturm and Collegiate}. The rows from top to bottom show feature map inversions from $5$ convolutional layers relu1\_2, relu2\_2, relu3\_2, relu4\_2 and relu5\_2 respectively. In every row, the columns from left to right show the inversion results of first five feature map respectively.

\begin{figure}[tbp]
	\centering
	\includegraphics[width=240pt, keepaspectratio]{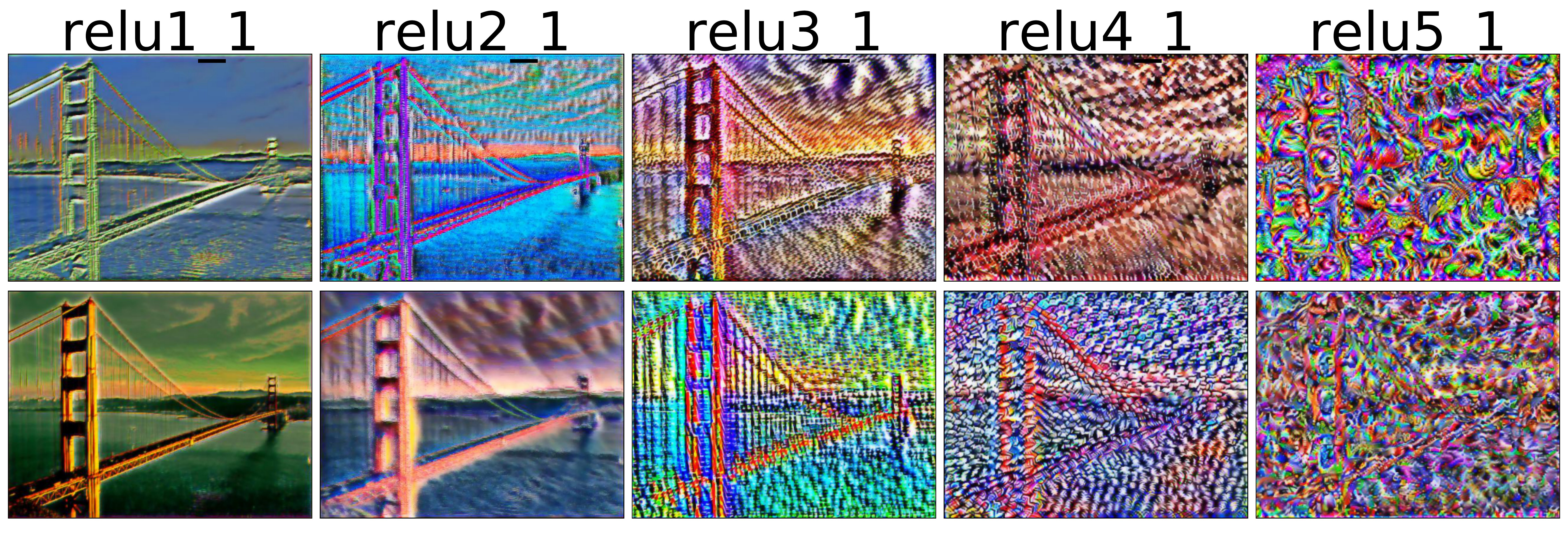}
	\caption{Randomly Modified Code Inversion. The columns from left to right show results at layers relu1\_1, relu2\_1,relu3\_1, relu4\_1 and relu5\_1. This figure is best viewed electronically in color with zoom.}
	\label{fig:random}
\end{figure}

Numerical results show that every filter extracts a specific texture. As Fig. \ref{fig:feature_map} shows, inversion results of different feature maps at different layers have distinct textures, while the corresponding inversion results in (a) and (b) have same textures including color and local structure. FMI for low layers such as relu1\_2 and relu2\_2 generates images whose color is monotonous and local structure is simple. As layers increase higher such as layer relu4\_2 and relu5\_2, the colours become plentiful and the local structures become more intricate. This character is reasonable because feature maps at higher layers can be considered as a non-linear combination of preceding feature maps. For example, feature maps at low layers represent low level semantic properties such as edge and corner, then posterior filters assemble different edge patterns and corner patterns to compose more intricate texture.

\begin{figure*}[tbp]
	\centering
	\includegraphics[scale=0.22]{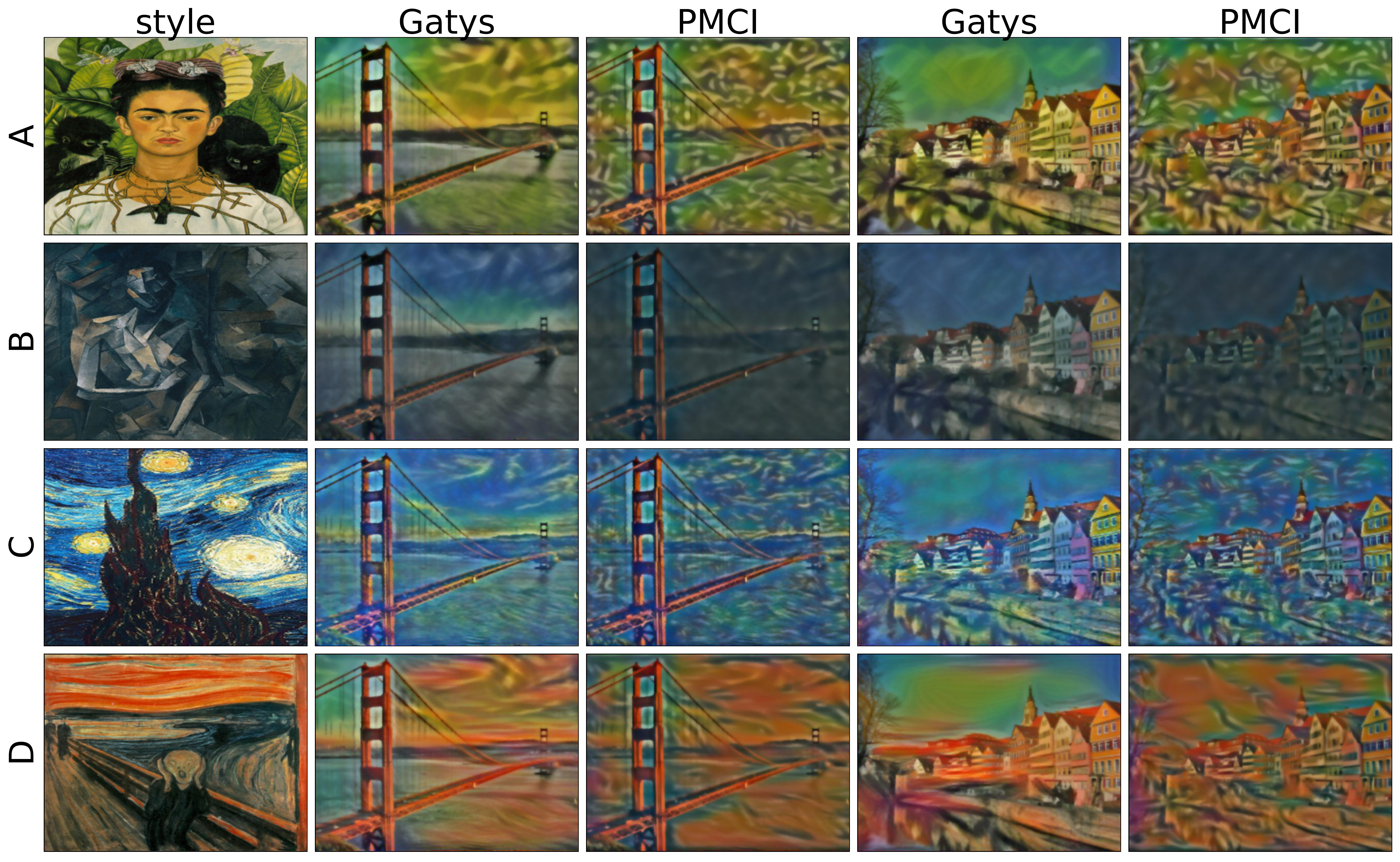}
\caption{Purposefully Modified Code Inversion (PMCI). From up to down, target style images are $\bm{A}$ \emph{Self Portrait with Necklace of Thorns} by Frida Kahlo in 1940, $\bm{B}$ \emph{Femme nue assise} by Pablo Picasso in 1909, $\bm{C}$ \emph{The Starry Night} by Vincent van Gogh in 1889 and $\bm{D}$ \emph{Der Schrei} by Edvard Munch in 1893. The third column and fifth column show PMCI results. The second column and the fourth column show styled images with the approach of Gatys \cite{Gatys_2016_CVPR}. To make all images alignment, we resize style target images.}
	\label{fig:style_transfer}
\end{figure*}

\subsection{Generating Images of Diverse Styles}
\label{sec:result_MCI}
Since every feature map represents a specific texture, we can change images' style by modifying the combination weights of hierarchical texture primitives randomly and purposefully. 

Fig. \ref{fig:random} shows the qualitative \textbf{Randomly Modified Code Inversion} results. We stochastically reallocate the sum of every feature map at layers relu1\_1, relu2\_1, relu3\_1, relu4\_1 and relu5\_1 respectively. For every layer we generate two random inversion results. Random modification changes the activation degree of activated neurons but keeps the unactivated ones invariant. Two generated images at same column have disparate texture. Compared with input images, color is the main difference at low layers relu1\_2, relu2\_2 and structure is the main difference at layers relu4\_1, relu5\_1, which supports the discovery of Sec. \ref{sec:result_FMI}. 

We find it is interesting that, inversion results at higher layers contain less content details and more specific texture. The reason is that the repeated block of textures at higher layers contains more intricate structure. However, image content is composed of many unique structural sub-images. When the structure of sub-images is different from the texture, the partial content information is destroyed. Finally, content of whole images becomes scarce and numerous intricate textures appear.

In addition, we also experiment with \textbf{Purposefully Modified Code inversion} (PMCI). The generated images in Fig. \ref{fig:style_transfer} combine the code of a target content image and the energy distribution of feature maps of a target style image. In particular, we pick 4 style images for our experiment: \emph{A Self Portrait with Necklace of Thorns}, \emph{Femme nue assise}, \emph{The Starry Night} and \emph{Der Schrei}. We use code at layer relu2\_2 as content term constraint and the sum of feature maps along channel axis at layers relu1\_1, relu2\_1, relu3\_1, relu4\_1, relu5\_1 as the term of style constraint. The first column shows the target style images. The third column and fifth column show PMCI results. We also show the styled images of Gaty \cite{Gatys_2016_CVPR} at the second column and the fourth column.

PMCI generates images similar to style target. We can intuitively find that the generated images at same row have similar styles. The similarity demonstrates that the combination weights of feature maps represent image style. Thus, we can determine whether two images are of same style according to energy distribution of feature maps. With these results, we provide some insights to understand why Gram matrix of feature maps \cite{Gatys_2016_CVPR} could represent image style. Like the sum of feature maps along channel axis, Gram matrix also guides the energy of every feature map of generated image.

\section{CONCLUSION}
\label{sec:conclusion}
We present a method to visualize which feature a filter captures from input image in CNNs by inversing the feature map of interest. By this technique, we demonstrate that every filter extracts a specific texture. The inversion result at higher layers contains more colors and more intricate structures. We propose two methods to generate images of diverse styles. The experimental results support our assumption that the style of an image is essentially a combination of texture primitives captured by filters in CNNs. In addition to generate images of diverse styles, we also provide an explanation about why Gram matrix \cite{Gatys_2016_CVPR} of feature maps could be a representation of image style. Since every filter extracts a specific texture, the combination weights of feature maps decide the image style. Like the sum of feature maps along channel axis, Gram matrix also guides the energy of every feature map of generated image.

\section{ACKNOWLEDGMENT}
\label{sec:acknowledgment}
This research is supported by National High Technology Research and Development Program of China(863, No. 2015AA015903), National Natural Science Foundation of China (NSFC, No. 61571102), the Fundamental Research Funds for the Central Universities (No. ZYGX2014Z003, No. ZYGX2015KYQD004) and a grant from the Ph.D. Programs Foundation of Ministry of Education of China (No. 20130185110010). We gratefully acknowledge the support of NVIDIA Corporation with the donation of Quadro K5200 GPU used for this research.  

\bibliographystyle{IEEE}
\bibliography{citation}
\label{sec:refs}
\end{document}